\documentclass[letterpaper, 10 pt, conference]{ieeeconf}  
\usepackage[utf8]{inputenc}

\usepackage{multirow}

\usepackage{graphicx} 
\usepackage{amsmath}
\usepackage{amssymb}
\usepackage[english]{babel}
\usepackage{color}
\usepackage{algorithm}
\usepackage{csquotes}
\usepackage{algpseudocode}
\DeclareMathOperator*{\argmax}{arg\,max}

\usepackage{float}
\usepackage{commath}
\usepackage{booktabs}
\usepackage{hyperref}
\usepackage{array}

\usepackage[
backend=biber,
style=ieee,
sorting=none,
url=false
]{biblatex}

\IEEEoverridecommandlockouts                              

\title{\LARGE \bf
Direct Random Search for Fine Tuning of Deep Reinforcement Learning Policies
}

\author{Sean Gillen, Asutay Ozmen, Katie Byl 
\thanks{*Sean Gillen, Asutay Ozmen, and Katie Byl are with the Department of Electrical and Computer Engineering. University of California, Santa Barbara CA 93106, USA 
\tt\small \url{sgillen@ucsb.edu},\tt\small \url{ozmen@ucsb.edu},\tt\small\url{katiebyl@ucsb.edu}}
\thanks{**Code hosted at: \url{github.com/sgillen/policy_refinement}}

}
\date{September 2021}

\begin{document}


\maketitle
\thispagestyle{empty}
\pagestyle{empty}


\begin{abstract}
Researchers have demonstrated that Deep Reinforcement Learning (DRL) is a powerful tool for finding policies that perform well on complex robotic systems. However, these policies are often unpredictable and can induce highly variable behavior when evaluated with only slightly different initial conditions. Training considerations constrain DRL algorithm designs in that most algorithms must use stochastic policies during training. The resulting policy used during deployment, however, can and frequently is a deterministic one that uses the Maximum Likelihood Action (MLA) at each step. In this work, we show that a direct random search is very effective at fine-tuning DRL policies by directly optimizing them using deterministic rollouts. We illustrate this across a large collection of reinforcement learning environments, using a wide variety of policies obtained from different algorithms. Our results show that this method yields more consistent and higher performing agents on the environments we tested. Furthermore, we demonstrate how this method can be used to extend our previous work on shrinking the dimensionality of the reachable state space of closed-loop systems run under Deep Neural Network (DNN) policies. 


\end{abstract}
\section{Introduction}

In recent years, researchers have leveraged Deep Reinforcement Learning (DRL) to solve a wide variety of continuous control problems. Examples include problems from the computer graphics community, in which DRL has been used for physics-based character animation~\cite{2018-TOG-deepMimic}, and a wide variety of complex robotic tasks. This paper focuses primarily on applications in robotics, which has seen an explosion of work in recent years \cite{lillicrap_continuous_2015} \cite{schulman2017proximal} \cite{haarnoja2018soft} \cite{Gillen2020CombiningDR}. Continuous control problems in the context of robotics include controlling a 47 degree-of-freedom (DOF) humanoid to navigate various obstacles~\cite{heess_emergence_2017}, dexterously manipulating objects with a 24 DOF robotic hand~\cite{openai_learning_2018}, training the quadrupedal ANYmal robot to recover from falls~\cite{lee_robust_2019}, and teaching the bipedal Cassie robot to navigate stairs blindly~\cite{siekmann2021blind}. These problems are all high dimensional, nonlinear, and underactuated, and they all involve complex contact sequences with the environments, which makes them very challenging for more traditional control design. Traditional model-based control techniques are still very effective---arguably, Boston Dynamics still represents the state-of-the-art for legged locomotion in robotics, for example. However, these approaches require hundreds of expert person hours to develop each new controller. DRL attempts to automate at least some aspects of this challenging controller development process. There are already examples of learned policies outperforming ones hand-designed by experts~\cite{hwangbo_learning_2019}, and with the ever-continued growth and availability of computational power, there is good reason to believe these learning methods will continue performing better and becoming easier to use.


But, of course, there are significant drawbacks to these model-free approaches. While Deep Neural Networks (DNNs) are very powerful, they also need to acquire a lot of data during training. This contributes to DRL being very sample inefficient, meaning that many interactions with the environment are required in order to find a good policy.  As a result, most training for robotic systems must be done in simulation, where the environment can be parallelized and run thousands of times faster than real time. Transfer learning is often required to adapt such policies so that they work for real-world hardware. Doing so effectively remains an important, open problem. Furthermore, modern DRL algorithms can be difficult to implement, as small implementation details can change performance dramatically~\cite{engstrom2020implementation}, which motivates our additional focus in reducing the observed variability in performance of closed-loop policies from DRL.


DRL policies are almost always stochastic in nature. During training almost all the common DRL algorithms either add exploration noise to the actions, or learn a probability distribution from which to sample at training time. This might be, for example, a simple Gaussian distribution, the more sophisticated Ornstein-Uhlenbeck correlated noise process in the case of Deep Deterministic Policy Gradient DDPG~\cite{lillicrap_continuous_2015}), or, in the case of DQN, a random selection of sub-optimal actions~\cite{mnih2015humanlevel}. However, when policies are deployed or evaluated, one typically uses a deterministic policy by taking what we will call the Maximum Likelihood Action (MLA). 

In \cite{Mania2018}, the authors show that a simple Augmented Random Search (ARS) over linear functions was competitive with deep reinforcement learning across a standard suite of benchmark tasks. Furthermore, this algorithm is simple and, in the cases the authors tested, around fifteen times more sample efficient than the best-performing DRL baseline. Despite these advantages, the simplicity of the policy class limits the environments to which it can currently be applied.




In this work, we show that a slightly modified version of this random search can be applied directly to DNNs for fine tuning, without any apparent loss in sample efficiency. The simplicity of this approach has several advantages. The first is that it does not appear to be very sensitive to hyper-parameter settings. We are able to use a single set of parameters for all the results obtained in this paper, across a dozen environments, and with most systems being tested for six different initial policies each obtained from a different DRL algorithm. Second, we avoid some of the previously mentioned problems stemming from the complexity and fragility of modern DRL algorithms. Finally, 
our data thus far indicate that we seem to achieve essentially the same sample efficiency seen in ARS, despite operating over much larger parameterizations.

We show that our proposed method of fine tuning leads to modest increases in reward and substantial improvements to consistency in performance for DRL agents across a large set of RL environments. In addition, we also show that we can also use this fine-tuning method to extend previous work of ours involving an extra dimensionality term in the reward~\cite{Gillen2020ExplicitFractal}.

The rest of this paper is laid out as follows. First, we introduce the problem statement, the algorithms and environments used, and implementation details for the training. We then present results obtained from using this policy refinement approach across a collection of continuous-control RL environments. We also perform some analysis on how often fall events occur for a benchmark bipedal walker where the existing DRL baselines are particularly prone to failure. After this, we present results of using this method to train with additional, dimensionality based reward terms in order to show that we are able to extend our previous work to DNN policy classes. Finally, we demonstrate the approach on a Panda arm simulation environment, where our approach leads to considerably smoother policies that avoid unwanted jitter, without any environment specific or algorithm specific tuning or reward shaping.  

\begin{table*}[!hbtp]
\label{tab:1}
\centering
\begin{tabular}{p{2cm} p{2cm}p{1.5cm}p{1.5cm}p{1.5cm}p{1.5cm}p{1.5cm}p{1.5cm}}
Environment &        &  A2C &                       PPO &                      DDPG &    TD3 & SAC &                       TQC \\
\bottomrule
\end{tabular}

\begin{tabular}{p{2cm}p{2cm}p{1.5cm}p{1.5cm}p{1.5cm}p{1.5cm}p{1.5cm}p{1.5cm}}
\toprule
\multirow{2}{*}{MountainCar}
&Baseline Return &  91 ± 0.2 &   88 ± 2.3 &  93 ± 0.0 &  93 ± 0.1 &  94 ± 1.3 &  \textbf{67 ± 43.8} \\
&Tuned Return    &  92 ± 0.1 &  96 ± 17.0 &  94 ± 0.4 &  94 ± 0.2 &  95 ± 1.1 &   \textbf{96 ± 0.9} \\
\end{tabular}

\begin{tabular}{p{2cm}p{2cm}p{1.5cm}p{1.5cm}p{1.5cm}p{1.5cm}p{1.5cm}p{1.5cm}}
\toprule
\multirow{2}{*}{LunarLander}
&Baseline Return &  61 ± 137.3 &  273 ± 30.5 &  216 ± 100.0 &  \textbf{205 ± 86.7} &  259 ± 67.8 &  279 ± 28.6 \\
&Tuned Return    &  160 ± 126.1 &  275 ± 32.4 &   249 ± 68.5 &  \textbf{257 ± 20.1} &   283 ± 18.1 &  286 ± 17.7 \\
\end{tabular}

\begin{tabular}{p{2cm}p{2cm}p{1.5cm}p{1.5cm}p{1.5cm}p{1.5cm}p{1.5cm}p{1.5cm}}
\toprule
\multirow{2}{*}{BoxWalker}
&Baseline Return &  \textbf{296 ± 27.0} &  \textbf{220 ± 122.4} &  \textbf{217 ± 127.4} &  \textbf{302 ± 65.1} &  \textbf{289 ± 66.0} &  \textbf{326 ± 58.2} \\
&TunedReturn &  \textbf{313 ± 0.7} &   \textbf{325 ± 0.7} &   \textbf{281 ± 54.1} &  \textbf{334 ± 0.6} &    \textbf{321 ± 1.0} &   \textbf{344 ± 0.3} \\
\bottomrule
\end{tabular}

\begin{tabular}{p{2cm}p{2cm}p{1.5cm}p{1.5cm}p{1.5cm}p{1.5cm}p{1.5cm}p{1.5cm}}
\multirow{2}{*}{BoxWalkerHard}
&Baseline Return &   99 ± 129.3 &  137 ± 119.4 &  N/A &  -92 ± 16.3 &  16 ± 104.2 &  238 ± 102.0 \\
&Tuned Return    &  109 ± 121.0 &  137 ± 119.7 &  N/A &   -23 ± 5.2 &   44 ± 86.4 &  242 ± 107.6 \\
\bottomrule
\end{tabular}


\begin{tabular}{p{2cm}p{2cm}p{1.5cm}p{1.5cm}p{1.5cm}p{1.5cm}p{1.5cm}p{1.5cm}}
\multirow{2}{*}{Walker2D}
&Baseline Return &   785 ± 389.2 &  2108 ± 16.0 &  \textbf{1432 ± 720.1} &  \textbf{2218 ± 194.6} &  \textbf{2290 ± 34.8} &  \textbf{2540 ± 557.6} \\
&Tuned Return &  913 ± 269.3 &  2250 ± 194.1 &   \textbf{1896 ± 375.7} &   \textbf{2411 ± 7.5} &   \textbf{2413 ± 13.6} &    \textbf{2812 ± 8.8} \\
\bottomrule
\end{tabular}

\begin{tabular}{p{2cm}p{2cm}p{1.5cm}p{1.5cm}p{1.5cm}p{1.5cm}p{1.5cm}p{1.5cm}}
\multirow{2}{*}{HalfCheetah}
&Baseline Return &  2109 ± 36.3 &  2938 ± 53.7 &  2064 ± 198.7 &  2820 ± 21.0 &  2792 ± 10.9 &  \textbf{3676 ± 16.7} \\
&Tuned Return    &  2211 ± 35.9 &  3000 ± 42.3 &  2264 ± 133.1 &  2928 ± 15.4 &   2883 ± 6.9 &  \textbf{3802 ± 11.9} \\
\bottomrule
\end{tabular}

\begin{tabular}{p{2cm}p{2cm}p{1.5cm}p{1.5cm}p{1.5cm}p{1.5cm}p{1.5cm}p{1.5cm}}
\multirow{2}{*}{Hopper}
&Baseline Return &   \textbf{834 ± 343.3} &  \textbf{2523 ± 383.5} &   \textbf{1179 ± 453.1} &   2681 ± 27.2 &  2602 ± 205.2 &  \textbf{2631 ± 329.7} \\
&Tuned Return &  \textbf{1643 ± 204.1} &    \textbf{2633 ± 91.0} &  \textbf{2379 ± 341.6} &  2749 ± 337.1 &   2706 ± 96.7 &   \textbf{2782 ± 20.7} \\
\bottomrule
\end{tabular}

\begin{tabular}{p{2cm}p{2cm}p{1.5cm}p{1.5cm}p{1.5cm}p{1.5cm}p{1.5cm}p{1.5cm}}
\multirow{2}{*}{Ant}
&Baseline Return &  2502 ± 25.4 &   2869 ± 72.7 &  2365 ± 212.5 &  \textbf{3268 ± 288.8} &  3096 ± 31.3 &  \textbf{3478 ± 24.0} \\
&Tuned Return    &  2679 ± 28.4 &  2897 ± 157.0 &   2424 ± 86.7 &   \textbf{3391 ± 24.8} &  3206 ± 18.0 &   \textbf{3654 ± 21.7} \\
\bottomrule
\end{tabular}


\caption{Average Return $\pm$ standard deviation before and after fine tuning}

\end{table*}
\begin{table}[hb]
\center
\small
\begin{tabular}{ llll }
\toprule
Env. & Algo. 
                  & Fail \% Before         &  Fail \% After \\ 
\midrule
\multirow{4}{*}{Walker}


& A2C   &       19.33   &       12.00 \\
& PPO   &       0.00    &       0.33 \\
& DDPG  &       42.67   &       4.00 \\
& TD3   &       12.33   &       1.67 \\
& SAC   &       2.33    &       0.67 \\
& TQC   &       5.67    &       0.00 \\

\bottomrule
\end{tabular}
\caption{\label{tab:falls} Measured early termination events before and after the fine tuning process \\
}
\end{table}

\section{Methods}

\subsection{Reinforcement Learning}

The goal of reinforcement learning is to train an agent acting in an environment to maximize some reward function. At every timestep $t \in \mathbb{Z}$, the agent receives the current state $s_{t} \in \mathbb{R}^{n}$, uses that to compute an action $a_{t} \in \mathbb{R}^{b}$, and then receives the next state $s_{t+1}$, which is used to calculate a reward $r : \mathbb{R}^{n} \times \mathbb{R}^{m} \times \mathbb{R}^{n} \rightarrow \mathbb{R}$. The objective is to find a policy  $\pi_{\theta}: \mathbb{R}^{n} \rightarrow \mathbb{R}^{m}$

\begin{equation} \argmax_{\theta} \mathop{\mathbb{E}}_{\eta}\left[ \sum_{t=0}^{T}r(s_{t}, a_{t}, s_{t+1}) \right] \end{equation}
where $\theta \in \mathbb{R}^{d}$ is a set that parameterizes the policy, and $\eta$ is a parameter representing the randomness in the environment. We will call the sum of rewards obtained during an episode a return. 

\subsection{Direct Policy Search}
\label{sec:search}

We start by noting that there are many names for what we are calling direct policy search, as it is at least 50 years old and has been rediscovered by a variety of different optimization communities. Algorithm~\ref{algo:brs} outlines our particular version of it. In essence the random search chooses 2n candidate policies at each step by adding zero-mean Gaussian noise to the current policy parameters. These candidate policies are used to perform rollouts, and the reward for each rollout is recorded. These rewards are then used in the update step for the policy.

In~\cite{Mania2018}, the authors show that this direct policy search is competitive with DRL. Specifically, they add a number of "tricks" to the basic algorithm and call their resulting approach the Augmented Random Search (ARS). We add our own set of tricks in this work. First, we keep ARS's update step, where the step size is divided by the standard deviation of returns obtained. Second, we maintain the normalization functions learned by the DRL algorithms we are tuning. This differs from algorithm to algorithm, but usually it involves normalization of the data using statistics of the observation that is seen during training, followed by a clipping operation. We also found that it is important to be careful with the random seeds used for rollouts, and so for each policy pair $\theta \pm \delta_{i}$ we ensured that the environment used the same seed. This was particularly important for environments with a wide distribution of initial conditions. Finally, we found performance was slightly improved by using a linear schedule for step size and exploration noise. 

One advantage of this method that we have found is that it is not very sensitive to hyper-parameters. For every result presented in this paper, we deliberately used the same parameters: 200 update steps, $n=64$, $\alpha=[0.02, .002]$, and $\sigma=[0.025, 0.0025]$, which were chosen using the parameters used in~\cite{Mania2018} as a starting point. Ignoring for a second that 200 update steps is in fact more than is necessary for most environments, this implies that our method takes 25600 rollouts to train. In simulation with parallel rollouts, this is completed in a matter of minutes using a Ryzen 3900x. For the Panda arm environments that we will discuss in more detail later, this would correspond to about 14 hours of real robot time, and we suspect this time could be brought down considerably by tuning the hyper-parameters specifically for sample efficiency. 

We also note that we ran experiments where we train only a subset of the neural network parameters, which would make the number of trainable parameters comparable to the linear policies used in~\cite{Mania2018}. During these experiments we found the results were slightly inferior to training on the entire network, and that the sample efficiency, measured by number of updates required to reach a given reward threshold, was almost exactly the same.

\begin{algorithm}
\caption{Direct Policy Search}\label{algo:brs}
\begin{algorithmic}
\Require Policy $\pi$ with trainable parameters $\theta$
\Require Hyper-parameters - $\alpha$ $\sigma$ $n$
\State Sample $\bf{\delta} = [\delta_{1}, ..., \delta_{n}]$ from $\mathcal{N}(0, \sigma)^{\text{n x }\abs{\theta}}$
\State $\theta^{*}  = [\theta - \delta_{1}, ..., \theta - \delta_{n}, \theta + \delta_{1}, ..., \theta + \delta_{n}] $
\For{$\theta_{i}$ in $\theta^{*}$}
    \State Do rollout with policy $\pi_{\theta_{i}}$, using the MLA
    \State Collect sum of rewards $R_{i}$. 
\EndFor
\State $ \theta^{+} = \theta + \frac{\alpha}{n \sigma_{R}}\sum_{i=0}^{n} (R_{i} - R_{i+n})\delta_{i} $ 
\end{algorithmic}
\end{algorithm}

\subsection{Environments}

\label{sec:envs}

We examine a number of popular benchmarking environments from the RL community. The environments all conform to the OpenAI Gym API introduced in~\cite{1606.01540}. For ease of reference, we will refer to each environment by the ID it has in the Gym registry. MountainCarContinuous-v0, LunarLandarContinuous-v2, BipedalWalker-v3, and BipedalWalkerHardCore-v3 are all standard continuous control environments included with the base Gym environments. To the best of our knowledge these environments are not meant to be physically realistic. We also study a collection of locomotion environments implemented in PyBullet. The locomotion environments were created by~\cite{gallouédec2021multigoal} and are maintained by the Bullet Physics team~\cite{coumans2020}. In this work we study  HalfCheetahBulletEnv-v0, HopperBulletEnv-v0, Walker2DBulletEnv-v0, and AntBulletEnv-v0. All of these environments are simulated legged robots. Agents take joint angles and velocities as input states, and compute joint torques as actions. The reward functions are designed to encourage agents to walk forward as fast as possible. It may be worth noting that these are inspired by OpenAI's popular Mujoco environments, though the Bullet versions are considerably heavier and impose more realistic torque limits, which makes them a bit more challenging for RL algorithms. In the second half of this paper, we study a set of environments based on a 7DOF Franka Emika Panda arm~\cite{gallouédec2021multigoal}. These environments are made difficult both by their complexity and the fact that they use a sparse reward structure. As an example of these aspects, consider the PandaPickAndPlace-v1 environment, in which the arm must pick up a block somewhere in its workplace and bring it to a randomized goal state. The agent recieves a reward of -1 everywhere except when the block has reached the goal state.


\subsection{Pre Trained Agents}

We use the Stable Baselines 3 Zoo \cite{rl-zoo3} \cite{stable-baselines3} for a collection of pretrained agents with tuned hyper parameters. The Zoo provides agents for Truncated Quantile Critics (TQC),  Soft Actor Critic (SAC), Proximal Policy Optimization (PPO), Asynchronous Actor Critic (A2C), Deep Deterministic Policy Gradients (DDPG), and Twin Delayed Deep Deterministic policy gradient (TD3) \cite{kuznetsov2020controlling} \cite{haarnoja2018soft} \cite{schulman2017proximal} \cite{mnih2016asynchronous} \cite{lillicrap_continuous_2015} \cite{fujimoto2018addressing}. In all examples, the policies are deep neural networks and the exact architecture has been tuned by the Zoo maintainers to have reasonable performance for each environment algorithm pair. We use these policies to initialize the values of $\theta$ in Algorithm~\ref{algo:brs}. 

\bgroup
\def\arraystretch{1.1}%
\begin{table*}[h]
\centering
\begin{tabular}{p{2cm} p{2cm}p{1.5cm}p{1.5cm}p{1.5cm}p{1.5cm}p{1.5cm}p{1.5cm}}
Environment &        &  A2C &                       PPO &                      DDPG &    TD3 & SAC &                       TQC \\
\bottomrule
\end{tabular}
\begin{tabular}{p{2cm}p{2cm}p{1.5cm}p{1.5cm}p{1.5cm}p{1.5cm}p{1.5cm}p{1.5cm}}
\toprule
\multirow{2}{*}{Walker2D}
&Baseline Dim.   &      2.55 ± 0.6 &      3.45 ± 0.4 &       5.54 ± 0.5 &      \textbf{6.09 ± 1.6} &       5.96 ± 1.6 &       \textbf{5.36 ± 0.5} \\
&Tuned  Dim.   &      1.21 ± 0.3 &      2.35 ± 0.2 &       3.82 ± 0.3 &       \textbf{3.72 ± 0.3} &       3.85 ± 0.5 &       \textbf{3.71 ± 0.2} \\
\cline{2-8}
&Baseline Return &   785 ± 389.2 &  2108 ± 16.0 &  1432 ± 720.1 &  \textbf{2218 ± 194.6} &  2290 ± 34.8 &   \textbf{2540 ± 557.6} \\
& Tuned  Return &    997 ± 2.2 &  2024 ± 10.1 &   1961 ± 12.5 &   \textbf{2152 ± 27.6}   &   2269 ± 13.3 &   \textbf{2562 ± 12.6} \\
\bottomrule
\end{tabular}

\begin{tabular}{p{2cm}p{2cm}p{1.5cm}p{1.5cm}p{1.5cm}p{1.5cm}p{1.5cm}p{1.5cm}}
\multirow{2}{*}{HalfCheetah}
&Baseline Dim.   &   3.19 ± 0.3 &   3.35 ± 0.2 &   \textbf{4.31 ± 0.4} &   \textbf{5.17 ± 0.3} &   4.83 ± 0.3 &   3.65 ± 0.2 \\
&Tuned  Dim.   &    2.4 ± 0.2 &   2.54 ± 0.2 &   \textbf{3.01 ± 0.3} &   \textbf{2.76 ± 0.3} &   3.46 ± 0.3 &   2.56 ± 0.2 \\
\cline{2-8}
&Baseline Return &  2109 ± 36.3 &  2938 ± 53.7 &  \textbf{2064 ± 198.7} &  2820 ± 21.0 &  2792 ± 10.9 &  3676 ± 16.7 \\
&Tuned  Return &  2137 ± 22.3 &  2778 ± 27.5 &  \textbf{2594 ± 41.9} &  2697 ± 13.1 &  2658 ± 12.1 &   3606 ± 7.2 \\
\bottomrule
\end{tabular}

\begin{tabular}{p{2cm}p{2cm}p{1.5cm}p{1.5cm}p{1.5cm}p{1.5cm}p{1.5cm}p{1.5cm}}
\multirow{2}{*}{Hopper}
&Baseline Dim.   &   2.85 ± 0.5 &    3.16 ± 0.5 &  3.67 ± 0.5 &   3.76 ± 0.4 &   \textbf{5.12 ± 0.3} &   \textbf{5.12 ± 0.3} \\
&Tuned  Dim.   &   2.24 ± 0.1 &    2.31 ± 0.2 &   3.12 ± 0.1  &   2.74 ± 0.1 &    \textbf{2.7 ± 0.2} &    \textbf{2.3 ± 0.1} \\
\cline{2-8}
&Baseline Return &  \textbf{834 ± 343.3} &  \textbf{2523 ± 383.5} &  \textbf{1179 ± 453.1} &   2681 ± 27.2 &  \textbf{2602 ± 205.2} &  \textbf{2631 ± 329.7} \\
&Tuned  Return &  \textbf{2072 ± 12.4} &   \textbf{2559 ± 26.0} &  \textbf{2641 ± 39.2} &   2763 ± 7.4 &   \textbf{2687 ± 8.1} &  \textbf{2547 ± 10.4} \\
\bottomrule
\end{tabular}

\begin{tabular}{p{2cm}p{2cm}p{1.5cm}p{1.5cm}p{1.5cm}p{1.5cm}p{1.5cm}p{1.5cm}}
\multirow{2}{*}{Ant}
&Baseline Dim.   &   2.65 ± 0.2 &   3.91 ± 0.6 &   7.14 ± 0.4 &    5.76 ± 0.2 &   7.17 ± 0.3 &    5.25 ± 0.3 \\
&Tuned  Dim.   &   2.15 ± 0.2 &   3.11 ± 0.1 &    6.87 ± 0.3 &    4.29 ± 0.4 &   3.35 ± 0.2 &    3.39 ± 0.2 \\
\cline{2-8}
&Baseline Return &  2502 ± 25.4 &   2869 ± 72.7 &  \textbf{2365 ± 212.5} &  3268 ± 288.8 &  3096 ± 31.3 &  3478 ± 24.0 \\
&Tuned  Return &  2527 ± 13.5 &  2817 ± 26.8 &  \textbf{2498 ± 42.9} &  3330 ± 100.1 &   2854 ± 8.0 &    3488 ± 3.4 \\
\bottomrule
\end{tabular}


\caption{Returns and Dimensionality after Fine Tuning with an extra dimensionality reward term}
\label{tab:mesh}

\end{table*}

\section{Results}

First we examine the results of using our direct policy search for policy fine-tuning of a large set of environments and initial policies, using the parameters from Section~\ref{sec:search}. We compare the mean and standard deviation of returns before and after our fine-tuning process. In both cases, the policies are evaluated deterministically by using the MLA at each step, the only randomness in the system is from the initial condition at the start of each episode, which is drawn from the same distribution seen during training. Each agent is evaluated with 100 Monte Carlo trials. 

The results are presented in Table~I. 
In almost all cases, we see at least a modest improvement to average return. Recalling that an even more fundamental goal in this work is to reduce variability, also note that many cases resulted in a substantial decrease in the variance of the return, as desired. This suggests that our fine-tuning process is effective both for squeezing extra performance out of a trained DRL agent and also for reducing the variability of those agents.

We also examine the robustness of these policies. We note that the baseline agents, even with no noise added and using the deterministic policy evaluation, will experience failure events from some particular initial conditions. Here we define failure as any "early termination" event from the environment. In the case of Walker2DBulletEnv-v0, the environment automatically terminates early if a non-foot link contacts the ground or if the simulation determines that a fall is imminent due to its center of mass location or body orientation. To test robustness, we sample 300 initial conditions and evaluate the policies both before and after our refinement step. We present the results from the Walker2D system because it had the highest failure rate across all baselines algorithms. We can see in Table~\ref{tab:falls} that DDPG, for example, failed in about 42\% of cases before the refinement process and in around 4\% afterwards. The other algorithms show improvement as well. In the case of TQC, we went from failing about 5\% of the time to not detecting any failure events during the 300 trials.

\section{Mesh Dimensions}

In previous work \cite{Saglam-RSS-14}, we introduced what we call a ``mesh dimension'' as an component to reward functions for reinforcement learning agents. Informally, agents typically operate in relatively high dimensional state spaces. However in practice they will often only move along a comparatively lower-dimensional manifold within that full space. That is, although motions are not completely synchronized over time, they demonstrate quite a bit of coordination among joints. By eye, such a gait-like coordination is often quite apparent. The mesh dimension attempts to identify this dimensionality reduction quantifiably. It estimates the dimensionality of the reachable state space of the closed-loop system, and, for those familiar with the term, it is very closely related to a ``fractal dimension''. 

In another line of prior work \cite{Gillen2020ExplicitFractal} \cite{gillen2021fractal}, we showed that ARS was able to train linear policies on environments which were modified to include this mesh dimension reward. This had a number of desirable qualities including finding very precise periodic gaits in some cases, and it improved robustness to push disturbances and sensor noise. In that work training used a lower and upper bound of the estimated dimensionality; in this work we train on the average of those two bounds. 

We experimented with several ways to incorporate this measure of dimensionality into the reward function, including both a linear and quadratic combination with the original reward. While these methods worked to some extent, they required fairly precise manual tuning of coefficients. Somewhat surprisingly, we found that simply taking the product of the original reward multiplied by the reciprocal of the dimension estimate $D$ was an effective reward that required no manual tuning:
\begin{equation}
\label{eq:Rr}
R^{r} = \frac{ \sum_{t=0}^{T}r(s_{t}, a_{t}, s_{t+1})}{D} .
\end{equation}

One caveat is that this only works for environments with positive rewards. For negative returns, however, as in the case of the Panda environments, we can simply take the product instead:
\begin{equation}
\label{eq:Rp}
 R^{p} = D \sum_{t=0}^{T}r(s_{t}, a_{t}, s_{t+1}) .
\end{equation}
We found that these rewards successfully gave the agents a signal to optimize, leading to significant reductions in dimensionality without any significant degradation in performance over the original reward.












\section{Mesh Dimension Results}

\subsection{Locomotion Environments}

First, we present results for fine-tuning with the post-processed reward from Equation~\ref{eq:Rp}. In this work we train DNNs on the more difficult Bullet environments. Results are shown in Table~III. All agents are evaluated deterministically here, using the MLA. Each entry in the Table for mean and standard deviation for the return and for the estimated dimensionality is calculated based on 100 Monte Carlo Trials. We see that the dimensionality (``Dim.'') in most environments is decreased quite drastically, particularly for the off-policy algorithms (DDPG, TD3). As before, this process also seems to decrease the variability of the return, perhaps even more reliably than without the dimension reward. We believe this shows that our previous results can be extended to DNNs, which greatly expands the scope of problems they can be applied to.

\subsection{Panda Arm Environments}


We present data here for a set of environments utilizing a Panda arm, introduced earlier in Section~\ref{sec:envs}. These environments present several challenging problems for an Emika Franka Panda arm. The PandaReach task is the most straightforward. Here, the goal is for the arm to reach a given point in task space. For PandaPush, the arm mush push a block along the floor to a desired location. In PandaSlide, the robot must grab a block and and bring it to a desired point on the ground. Finally, PandaPickAndPlace requires the arm to pick up a block and keep its grip on it while attempting to reach a point in space. We found that merely fine-tuning the action network with our random search did not improve performance significantly, though to be fair, none of the algorithms in the Zoo are able to solve this environment without Hindsight Experience Replay (HER)~\cite{andrychowicz2018hindsight}.

\begin{table}[!ht]
\begin{tabular}{lllll}
\toprule
          Environment & Base Dim. & Our  Dim. & Base. Return & Our  Return \\
\midrule
        PandaReach &    2.73 ± 0.7 &    2.28 ± 0.5 &        -2 ± 0.6 &        -1 ± 0.7 \\
 Pick\&Place &    1.63 ± 0.3 &    1.61 ± 0.5 &        -6 ± 2.6 &      -11 ± 13.3 \\
         PandaPush &    1.91 ± 0.5 &    1.68 ± 0.3 &        -6 ± 2.7 &        -7 ± 3.0 \\
        PandaSlide &    1.89 ± 0.4 &    1.53 ± 0.3 &       -22 ± 7.1 &      -41 ± 12.4 \\
\bottomrule
\end{tabular}
\caption{Dimensionality and Returns before and after fine tuning for the panda environments}
\end{table}

We did however also apply our dimensionality reward signal to this environment using our fine tuning process. We show the resulting dimensionality, and the returns in terms of the original reward function are shown in Table~IV. We observed a modest decrease in the dimensionality, accompanied by some decrease in the original return. Again, this decrease in reward is not unexpected, as we are after all trained on modified reward function. In addition to this, despite the Table data that suggests perhaps only a small change in behavior was observed, we noticed a significant beneficial change in the qualitative behavior of the robot. Figure~\ref{fig:panda} shows a stark example of this. In the PandaReach environment, the baseline agent is able to get its end effector into the target region, however it exhibits undesirable shaking behavior which the reward function does not punish. Our agent is able to achieve the same effect with a smooth motion. Note that both agents received exactly the same reward for the episodes we show,i.e. that despite the jittering deviations in the end effector, it remain in the goal region. 

\begin{figure}[!htb]
    \centering
    \includegraphics[width=\linewidth]{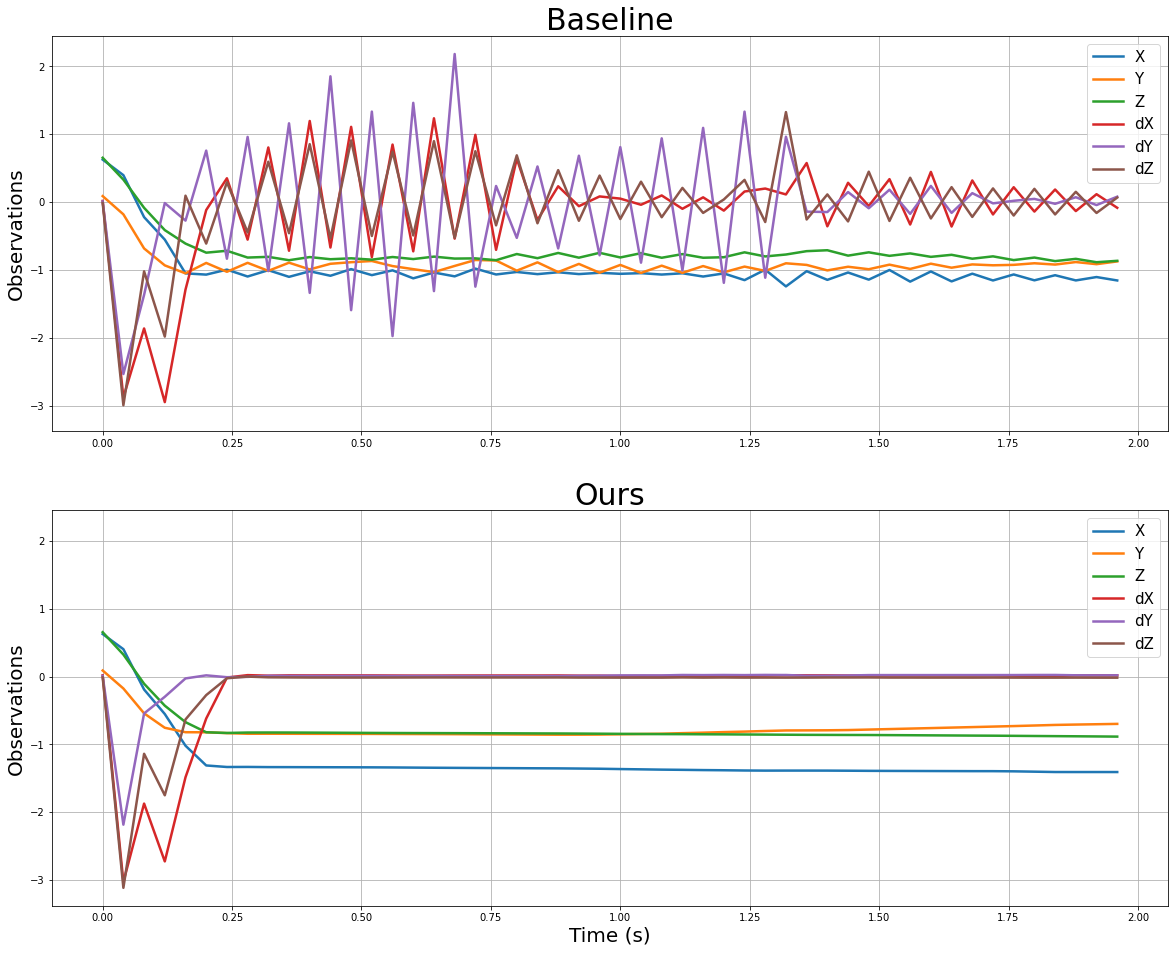}
    \caption{End effector positions and velocities for a policy roll out on PandaReach before and after fine tuning with the mesh dimension reward}
    \label{fig:panda}
\end{figure}
\egroup



\section{Discussion and Related Work}


It's worth discussing alternatives to our method for fine-tuning. The most similar work to ours that we have found is~\cite{pmlr-v100-xie20a}. There, the authors take a similar approach in that they decouple the algorithms used for exploration versus exploitation. We agree with their conclusion that this decoupling brings advantages on its own, regardless of the methods used for the fine-tuning exploitation. In some sense their work is doing the opposite of our approach, however, in that after using a gradient-free evolutionary strategy for exploration, it then uses DRL for exploitation (rather than for exploration). While the approaches are quite distinct, they are also in fact likely compatible, in that they could actually be combined. It is easy to imagine a pipeline using their gradient-free method for broad exploration, followed by DRL for initial exploitation, with our random search added for the final fine-tuning stage of an algorithm. 

There are also many small tricks and improvements found in DRL algorithms that aim to achieve similar results to what we've shown. One example is to decrease the SGD/Adam/RMSProp step size as training goes on. This is an effective method, and indeed our own method uses a linear schedule for the step size. However, the algorithms we are using as a baseline were already using this approach as well, and we still saw improvements in performance with the additional of our fine tuning process. 

Entropy regularization / penalties are another toolset available, which can also be put on a schedule. These can encourage an agent to use a wide distribution of actions initially and then gradually narrow this down as training continues. Again though, most of the algorithms used as a baseline (PPO, SAC, TQC) have some form of this already, and our method is still able to improve on them. 

We could try curriculum learning, meaning that the reward function could change by design as training goes on. We believe this is likely most effective when one has a lot of domain knowledge of the task, and when being applied to tasks that are too difficult for the algorithm to learn initially. For example, the authors of~\cite{xie2021policy} use this approach when controlling Cassie. This approach works well for them because they are able to engineer a reward that led to the desired behavior.

\section{Conclusion}

We have presented a method that can fine tune policies obtained from DRL algorithms by optimizing directly using the MLA. We showed that performance compared to a baseline was improved considerably on a large set of standard benchmarking tasks. Of more particular note, the variability of episode returns was decreased significantly on many of the environments we tested as well. For the system on which we also quantified failure rates (i.e., for the biped Walker), this lower variability was also accompanied by significantly fewer early termination events compared to the baseline. We hypothesize that this increased robustness is, quite plausibly, due to the dimensionality reduction. (That hypothesis is in fact why we performed these experiments, of course.) However, any conclusions on correspondence remain a topic for further investigation. 

We also showed that this method allows us to expand our previous work on adding dimensionality metrics to the reward function of RL agents to DNNs as well, which greatly expands the scope of problems it can be applied to. We demonstrated this on a set of locomotion environments and also on a challenging set of Panda arm environments with sparse reward structures. We showed that for the case of the Panda our approach achieved significantly less jitter, and arguably more visually pleasing (and mechanically desirable)
motion than the baseline, without any environment specific reward shaping, or manual adjustment of any parameters. 

We believe versatility and simplicity are major strengths of this approach. Policies obtained from any kind of DRL algorithm can be tuned in this way, and the method seems to require very little manual tweaking. The potential applications for this method are broad. Engineers designing robots which are public facing or that interact with humans may find it useful to employ policies that make their robots motions smoothing and thereby easier for humans to predict. There are applications outside of robotics as well. Physics-based character animation may also benefit from more consistently behaving policies, and DRL is also popular for video game AI, which is another area where the improved consistency of this method may prove desirable. 

Finally, we will end with a discussion on the broader impacts and future directions of this work. DRL has been an exciting and promising paradigm for robotic control for some time now, but it has yet to be widely adopted by industry. This is largely because it is difficult to trust a DNN controller, and deploying a poorly understood controller can be expensive and dangerous. By itself, we think the fine tuning method we've introduced can help make DRL policies more effective and reliable, however we also think that the lower dimensional policies can unlock even more tools to aid with this. With lower dimensional polices, we open the possibility to develop methods to perform numerical estimates of a variety of controls-based metrics, such as rates of contraction (Lyapunov exponents), identification of dangerous regions in state space (outside a stochastic separatrix for a basin of attraction), and/or expected (conservative) distributions of failure rate. All of these are promising directions towards safer and more reliable DRL based control, and we anticipate that our method brings us closer to realizing them for useful, real world, robotic systems.

\medskip

\printbibliography

\end{document}